\documentclass{esannV2}
\usepackage[dvips]{graphicx}
\usepackage[utf8]{inputenc}
\usepackage{amssymb,amsmath,array}
\usepackage{colortbl}
\usepackage{xcolor}
\usepackage{multirow}
\usepackage[normalem]{ulem}
\usepackage{afterpage}

\usepackage{lscape}     % rotate box 

\definecolor{colorblue}{rgb}{0.29, 0.59, 0.82}
\definecolor{colorred}{rgb}{0.9, 0.17, 0.31}
\definecolor{colorred}{rgb}{0.9, 0.17, 0.31}
\definecolor{coolgrey}{rgb}{0.55, 0.57, 0.67}
\newcommand{\dcb}{\cellcolor{colorblue!20}}

\newcommand{\dch}{\rowcolor{coolgrey!30}}
\newcommand{\dcf}{\rowcolor{coolgrey!20}}

\newcommand{\mr}[1]{\multirow{2}{*}{#1}}
\newcommand{\mrm}[1]{\multirow{-2}{*}{#1}}

\newcolumntype{"}{@{\hskip\tabcolsep\vrule width 1.5pt\hskip\tabcolsep}}
\makeatletter
\newcommand{\thickhline}{%
    \noalign {\ifnum 0=`}\fi \hrule height 1pt
    \futurelet \reserved@a \@xhline
}

\begin{document}
\title{A Systematic Assessment of Deep Learning Models for Molecule Generation}
\author{Davide Rigoni\textsuperscript{1,2}, Nicolò Navarin\textsuperscript{1} and Alessandro Sperduti\textsuperscript{1}
\thanks{The authors acknowledge the HPC resources of the Department of Mathematics, University of Padua, made available for conducting the research reported in this paper.}
%
% Optional short acknowledgment: remove next line if non-needed
% \thanks{1- drigoni@math.unipd.it}
\vspace{.3cm}\\
\textsuperscript{1}University of Padua - Department of Mathematics ``Tullio Levi-Civita''\\
via Trieste 63, 35121, Padua - Italy\\
\textsuperscript{2}Fondazione Bruno Kessler, via Sommarive 18, 38123 Trento, Italy
}
%***********************************************************************
% END OF AUTHORS INFORMATION AREA
%***********************************************************************

\maketitle

\begin{abstract}
In recent years the scientific community has devoted much effort in the development of deep learning models for the generation of new molecules with desirable properties (i.e. drugs).
This has produced many proposals in literature.
However, a systematic comparison among the different VAE methods is still missing.
For this reason, we propose an extensive testbed for the evaluation of generative models for drug discovery, and we present the results obtained by many of the models proposed in literature.
\end{abstract}

\section{Introduction}
The chemical space is so vast that, with the computational resources available nowadays, its complete exploration is impossible.
For this reason, in recent years the scientific community has devoted much effort in the study of deep learning models 
that are capable of generating candidate molecules that are likely to exhibit some pre-specified properties, 
%that allow us to concentrate our research
allowing researchers to focus just on a small part of this chemical space.
These methods can be used, for instance, in the process leading to the discovery of 
%novel 
molecules that can become new drugs.
%A very important exploration of this large space is that which seeks to find important molecules that can become new drugs.
Thanks to the development of increasingly effective deep learning models and the presence of large data sets, in recent years promising results have been achieved.

Starting from the model in~\cite{gomez2018automatic}, many other works have been proposed \cite{DBLP:conf/icml/KusnerPH17, DBLP:conf/iclr/DaiTDSS18, DBLP:conf/icml/JinBJ18, de2018molgan, DBLP:conf/nips/MaCX18, DBLP:conf/nips/LiuABG18}.
%carried out which have led to a continuous improvement of the state-of-the-art.
However, often different works use different metrics to evaluate their models, making a fair and objective comparison among models difficult.
For this reason we present a systematic approach to evaluate models for drug generation,
%to the generation of drugs,
using a precise set of metrics that aims to detect the various nuances that exist among the models.
In this way it is possible to fairly evaluate the models based on the %objectives
statistics and
properties of the molecules they generate.
This comparison is not possible just referring to the original works in literature.

In this work we focus mainly on VAE models, reporting the reader to~\cite{brown2019guacamol} for a comparison considering other types of models.
%in which models that apply genetic algorithms are also examined.}
%, which cannot be done nowadays in the State-of-the-Art.
% ------------------ before cutting 
%This paper is organized as follows.
%In Section~\ref{sec:models} the main models in literature are described.
%Section~\ref{sec:processesandmetrics} describes our proposed evaluation procedure and the metrics that will be considered. 
%In Section~\ref{sec:results}, we present the experimental evaluation of the models presented in %Section~\ref{sec:models}.Section~\ref{sec:conclusions} concludes the paper.

\section{Models for Molecule Generation}
\label{sec:models}
State-of-the-art models for drug generation are based on Variational Autoencoders (VAEs)~\cite{diederik2014auto} or on Generative Adversarial Networks (GANs)~\cite{arjovsky2017wasserstein}. 
The main idea %is to learn an approximation of the function that generated the training data, and then to use it to generate new data. In order to do this, 
of these approaches is to learn an embedding in a vector space (latent space) of the input data that aims to capture their properties and relations. Samples from the latent space are then used to generate 
%From this latent space we can sample new vectors that can be decoded in to
new data that are supposed to exhibit the properties of interest.
VAEs use an \emph{encoder} to encode an input molecule in the latent space and a \emph{decoder} to reconstruct the molecule corresponding to a point in the latent space.
%in the corresponding molecule. 
The GAN architecture is composed of a \emph{discriminator} and a \emph{generator}, both implemented by neural networks. 
The \emph{generator} performs the same function as the decoder in VAE: starting from a point in the latent space, in this case sampled from a standard normal distribution, it generates the corresponding molecule. 
The \emph{discriminator} aims to distinguish the input molecules that have been generated by the generator, from the ones in the training set.
The generator is 
%not trained directly and instead is 
trained only via the discriminator model, that provides its loss function.
%Specifically, the discriminator is learned to provide the loss function for the generator. 
%\todo{Rimossa questa frase sottostante.Vedi commenti latex}
%The two models compete in a two-player game, where simultaneous improvements are made to both generator and discriminator. % models that compete. 
Thanks to an ad-hoc optimization process~\cite{gomez2018automatic, DBLP:conf/icml/KusnerPH17, DBLP:conf/iclr/DaiTDSS18, DBLP:conf/icml/JinBJ18, DBLP:conf/nips/LiuABG18} or by introducing additional terms in the loss function~\cite{  de2018molgan, DBLP:conf/nips/MaCX18},
specific properties of the generated molecules can be optimized.
%Usually the optimization process is not included in the model that generates graphs, but it is performed by a second model with the purpose of searching the best graph in the latent space~\cite{gomez2018automatic, DBLP:conf/icml/KusnerPH17, DBLP:conf/iclr/DaiTDSS18, DBLP:conf/icml/JinBJ18}.
%However in some models generation and optimization are jointly considered~\cite{de2018molgan, DBLP:conf/nips/MaCX18}.
In this work, we only focus on comparing the generation capabilities of the different models, leaving the assessment of property optimization techniques as future work. 
%We thus decided to postpone the study of the different approaches for optimizing molecule properties in a future work.

A brief description of the models we consider follows.
\noindent The Character VAE~\cite{gomez2018automatic} exploits, in input and output, %receives in input the uses the 
SMILES~\cite{weininger1988smiles} %, weininger1989smiles, weininger1990smiles} 
strings describing the structure of the molecule. %, for both its input and output.
%and encode it in the latent space while the decoder uses a point in the latent space to reconstruct the SMILES representation of the molecule.
The encoder uses a convolutional network and the decoder uses a gated recurrent unit (GRU).
Since every molecule can be represented as a SMILES string but not vice-versa, this approach generates many strings that do not correspond to actual molecules. 
%To solve this problem, 
Grammar VAE~\cite{DBLP:conf/icml/KusnerPH17} 
%improves Character VAE 
attempts to enforce the syntactic validity of SMILES strings by introducing a context-free grammar to direct the generation process.
%INSERIRE SE AVANZA SPAZIO In this way encoder and decoder learn the production rules that generated the SMILES strings in input, filtering in output at each GRU time step the production rules that are not possible according to the grammar.
%
%This model attempts to enforce the syntactic validity of SMILES representations.
Syntax Directed VAE~\cite{DBLP:conf/iclr/DaiTDSS18} improves Grammar VAE using a more expressive grammar (a variant of the attribute grammar) which aims to generate strings that not only are syntactically valid, but also semantically reasonable.
Junction Tree VAE~\cite{DBLP:conf/icml/JinBJ18}%, unlike previously described models,
%does not exploit the SMILES representation of molecules, but it 
~represents molecules using graphs, composed of chemical substructures that are extracted from the training set.
%First, the algorithm looks in the training set for a list of substructures %that are easy 
%to be adopted as building blocks for the decomposition and reconstruction of the molecule.
%Its main idea is to 
%The model generates 
New molecular graphs are obtained by first
generating a tree-structured scaffold formed by substructures (the junction tree), and then combining the substructures together using a graph message passing network.
%in order to build the molecule.
%The advantage of this model is to reconstruct the molecule piece by piece %considering how the substructures of the list are joined, 
%rather than directly generating all the atoms.
%
Regularized Graph VAE~\cite{DBLP:conf/nips/MaCX18}  %uses the molecule graph representation as well.
%It 
casts the molecule generation problem as a constrained optimization problem, where %, since in order for a molecule to be valid, it must respect a series of 
 chemical constraints are encoded in the VAE loss function.
%existing between the bonds and the nodes. 
%Therefore it is possible to see the problem of the generation of molecules as a constrained optimization problem.
%subject to real constraints. 
%Regularized Graph VAE transforms the constrained optimization problem into a regularized unconstrained one, applying a generalization of the Lagrangian function in order 
%Authors transform the optimization problem to move all the explicit constraints into the VAE loss function. 
The encoder and decoder are implemented with a convolutional and deconvolutional networks, respectively.
%The decoder generates directly  atoms and bonds (including their types).
Constrained Graph VAE~\cite{DBLP:conf/nips/LiuABG18} %presents the best results in literature.
%The idea behind this model is to use a 
%latent space that encodes atoms rather than whole molecules.
%differently from previously described methods, 
encodes in the latent space single atoms rather than whole molecules.
%For this reason, 
%In order 
To generate a molecule, first the model samples several nodes in the latent space and assigns them an atom type using a linear classifier; then it connects them using a (constrained) breadth first algorithm.
%that tries to respect all the chemical rules. 
Both the encoder and decoder are implemented by a gated graph sequence neural network (GGNN)~\cite{DBLP:journals/corr/LiTBZ15}.
MolGan~\cite{de2018molgan}, based on generative adversarial networks, % and it works with the molecule graph representation.
%It 
learns via reinforcement learning to directly reconstruct, by a multi-layer perceptron, the molecular graph by predicting directly the atoms type, and the existence of bonds (and their types). 
%Although this model has good results, it suffers from the model collapse problem e.g tends to always generate the same molecules.
%\todo{Rimossa questa frase sottostante.Vedi commenti latex}
%In Section~\ref{sec:results} we experimentally compare all the above described models.

\section{Evaluation Processes and Metrics}
%\section{Processes and Metrics}
\label{sec:processesandmetrics}
We consider two datasets of molecules: QM9~\cite{%ruddigkeit2012enumeration,
ramakrishnan2014quantum}, composed by about 134,000 organic molecules with a maximum of 9 atoms,
%with 108,000 organic molecules which contains molecules formed by a maximum of 9 atoms, 
%and another with 
and  ZINC~\cite{irwin2005zinc}, composed by 250,000 drug-like molecules with up to 38 atoms.
%extracted at random from the ZINC~\cite{irwin2005zinc} database which contains more complex molecules up to 38 atoms.
We fix the training and test splits for each dataset, that we release together with the code\footnote{https://github.com/drigoni/ComparisonsDGM.}. We choose a set of metrics trying to capture the strengths and weaknesses of the models (and to limit the required  computational efforts).
%considering the amount of computational times that all the models requires.
These are divided into two main categories: those that evaluate the generated molecules based on chemical properties
%of the molecules generated, 
and those that use only the information about their structure.
In the latter we find:  \emph{Reconstruction} that, given an input molecule and a set of generated molecules, %generated from its latent space representation,
computes the percentage of generated molecules that are equal to the one in input;
\emph{Validity} that, given a set of generated molecules, represents the percentage of them that is \emph{valid}, i.e. that represent actual molecules;% (according to the laws of chemistry);
\emph{Novelty} that represents the percentage of new  generated molecules, i.e. not in the training set;
\emph{Uniqueness} that represents (in percentage) the ability of the model to generate different molecules in output, and is computed as the size of the unique set of valid generated molecules divided the total number of valid generated molecules;
\emph{Diversity} that measures how much the generated molecules are different from those in the training set. This is a heuristic that uses randomly selected substructures present in the molecules.
\noindent The metrics used to measure the properties exhibited by the generated molecules are: 
\emph{Natural Product (NP)} which indicates how much the generated molecules structural space is similar to that covered by natural products~\cite{ertl2008natural};
\emph{Solubility 
(Sol.)}
%(logP)} 
which indicates how much a molecule is soluble in water;
\emph{Synthetic Accessibility Score (SAS)} which represents how easy~(0) or difficult~(100) it is to synthesize a molecule; \emph{Quantitative Estimation Drug-likeness (QED)} which indicates in percentage how likely it is that the molecule is a good candidate to become a drug. %This metric is related to the previous two.

The process used to generate the molecules on which to calculate the \emph{Reconstruction} metric consists of encoding each of the molecules in the test set 20 times (obtaining 20 slightly different representations), and decoding each of these points only once.
This process was chosen because both the encoder and the decoder always contain a probabilistic component and in this way we estimate the model's ability to reconstruct the molecule considering both factors.
\emph{GAN} model cannot compute the reconstruction metric since it cannot generate the latent space representation of a molecule.
%by encoding inside the molecules of the training set but uses the generator to generate new molecules starting from the latent space \(\sim \mathcal{N}(\mu=0,\,\sigma^{2}=1)\), it is not possible to find the encoding of a molecule in the training set. For this reason % with this model.
Since we are interested in the generation of new molecules, the other metrics are computed by another process that consists of directly sampling 20,000 points from the standard normal distribution and decoding each point only once.
%The authors of the reproduced models have always used a different or unbalanced processes which focus more on the encoder or the decoder part.
%\todo{Davide: Ho commentato la parte in cui discuti l'implementazione, secondo me non essenziale, si può mettere nel readme del repository. Ho messo una frase negli esperimenti in cui diciamo che nel nostro lavoro abbiamo trovato alcuni problemi nei vari software, e per questo motivo alcuni risultati non sono in linea con gli articoli originali. Così risparmiamo mezza pagina. Va detto però che abbiamo usato in genere gli iperparametri riportati nei paper, e va detto dove li abbiamo cambiati e perchè (tipo se su zinc non erano riportati, abbiamo raddoppiato le epoche)}
%%----- last changes
%For each model we adopted the values for the hyper-parameters used by the authors, where available. 
%For Graph VAE and Regularized Graph VAE the hyperparameters for the ZINC dataset were not specified. We thus decided to keep the same values provided for QM9 and, after preliminary results, to double the number of epochs.
Since the considered models require a high computational load, we adopted the hyperparameters values  reported in the original papers, when available. 
For Graph VAE and Regularized Graph VAE the hyperparameters for the ZINC dataset were not specified. 
We thus decided to keep the same values provided for QM9 and, after preliminary results, to double the number of epochs.

\section{Results}
\label{sec:results}
\afterpage{
\begin{landscape}
\begin{table}[t]
%\label{tab:results}
\scriptsize
\centering
\begin{tabular}{| l | c | c | c | c | c " c | c | c | c | c | c |}
\hline
\dch{}\textbf{Model trained on QM9}      &$\uparrow$\textbf{\%Rec.}    &$\uparrow$\textbf{\%Val.}&\textbf{$\uparrow$\%Nov.}&\textbf{$\uparrow$\%Uniq.}   &\textbf{$\uparrow$\%Div.} &$\uparrow$\textbf{\%NP}  &$\uparrow$\textbf{\%Sol.} &$\downarrow$\textbf{\%SAS} &$\uparrow$\textbf{\%QED}    \\
\hline
%   model                           reconstruction      validity            novelty         uniqueness          diversity          NP              Sol             SAS             QED
\mr{Character VAE}                  &2.99               &6.41               &99.38          &\mr{92.27}         &98.03              &81.92          &32.14          &43.48          &30.30              \\
                                    &\(\pm\)17.02       &\(\pm\)24.48       &\(\pm\)7.88    &                   &\(\pm\)9.25        &\(\pm\)11.40   &\(\pm\)25.13   &\(\pm\)30.32   &\(\pm\)15.95      \\
\hline
\mr{Grammar VAE}                    &\dcb{}58.54        &4.45               &94.22          &\mr{83.22}         &\dcb{}98.88        &79.91          &28.40          &33.75          &31.91              \\
                                    &\(\pm\)49.27       &\(\pm\)20.73       &\(\pm\)23.33   &                   &\(\pm\)7.71        &\(\pm\)14.60   &\(\pm\)20.66   &\(\pm\)31.15   &\(\pm\)12.05       \\
\hline
\mr{Syntax Directed VAE}            &52.54              &15.00              &\dcb{}100.00   &\dcb{}             &97.66              &82.60          &26.99          &22.92          &35.03              \\
                                    &\(\pm\)49.94       &\(\pm\)35.71       &\(\pm\)0       &\dcb{}\mrm{100.00} &\(\pm\)4.90        &\(\pm\)14.67   &\(\pm\)22.11   &\(\pm\)35.15   &\(\pm\)11.18       \\
\hline
\mr{Graph VAE*}                      &0.60               &89.06              &42.75          &\mr{85.74}        &66.94               &94.96          &37.28          &32.11          &48.34              \\
                                    &\(\pm\)7.72        &\(\pm\)31.22       &\(\pm\)49.47   &                   &\(\pm\)28.94       &\(\pm\)10.61   &\(\pm\)13.54   &\(\pm\)23.51   &\(\pm\)7.67        \\
\hline
\mr{Regularized GVAE*}              &0.66               &87.71              &41.26          &\mr{83.13}         &63.00              &\dcb{}96.32   &\dcb{}37.85    &28.89          &\dcb{}48.81              \\
                                    &\(\pm\)8.09        &\(\pm\)32.83       &\(\pm\)49.23   &                   &\(\pm\)27.91       &\(\pm\)9.00    &\(\pm\)13.24   &\(\pm\)23.40   &\(\pm\)7.14        \\
%\hline
%\mr{Regularized GVAE full*}          &0.57               &86.08              &42.43          &\mr{86.49}         &59.78              &\dcb{}96.42    &36.89          &29.79          &48.53              \\
%                                    &\(\pm\)7.50        &\(\pm\)34.61       &\(\pm\)49.42   &                   &\(\pm\)31.45       &\(\pm\)9.99    &\(\pm\)13.29   &\(\pm\)23.23   &\(\pm\)7.153        \\
\hline
\mr{Junction Tree VAE}              &53.88              &99.95              &91.14          &\mr{90.27}         &57.60              &91.46          &27.05          &18.97          &46.15              \\
                                    &\(\pm\)49.85       &\(\pm\)2.24        &\(\pm\)28.36   &                   &\(\pm\)30.75       &\(\pm\)15.23   &\(\pm\)13.59   &\(\pm\)20.70   &\(\pm\)7.88        \\
\hline
\mr{Constrained GVAE*}              &33.86              &\dcb{}100.00       &92.82          &\mr{98.86}         &79.13              &93.05          &27.96          &\dcb{}13.81    &46.78             \\
                                        &\(\pm\)47.32       &\(\pm\)0           &\(\pm\)25.82   &                   &\(\pm\)21.61       &\(\pm\)12.26   &\(\pm\)13.36   &\(\pm\)19.20   &\(\pm\)21.30        \\
\hline
\mr{MolGAN*}                        &                   &76.74              &56.22          &\mr{20.00}         &61.11              &96.27          &31.62              &31.09          &48.39        \\
                                    &\mrm{NA}           &\(\pm\)42.25       &\(\pm\)49.61   &                   &\(\pm\)35.94       &\(\pm\)41.30   &\(\pm\)17.13       &\(\pm\)21.28   &\(\pm\)14.58        \\
\thickhline
\dcf{}                              &\multicolumn{4}{c}{}                                                               &               & 88.52         & 27.91             & 21.86         & 46.12             \\
\dcf{}\mrm{Properties' Scores for Dataset QM9}             &\multicolumn{4}{c}{}                                                               &               &\(\pm\)17.75   &\(\pm\)13.76       &\(\pm\)22.88   &\(\pm\)7.76        \\
\hline
\end{tabular}
\\\vspace*{0.1cm}
\begin{tabular}{| l | c | c | c | c | c " c | c | c | c | c |}
\hline
\dch{}\textbf{Model trained on ZINC}      &$\uparrow$\textbf{\%Rec.}    &$\uparrow$\textbf{\%Val.}&$\uparrow$\textbf{\%Nov.}    &$\uparrow$\textbf{\%Uniq.}   &$\uparrow$\textbf{\%Div.} &$\uparrow$\textbf{\%NP}  &$\uparrow$\textbf{\%Sol.}&$\downarrow$\textbf{\%SAS} &$\uparrow$\textbf{\%QED}    \\
\hline
%   model                               reconstruction      validity            novelty             uniqueness               diversity       NP              Sol             SAS                 QED
\mr{Character VAE*}                      &25.28              &0.93               &\dcb{}100.00       &\mr{91.40}             &98.19          &80.82          &29.60              &31.11          &38.70              \\
                                        &\(\pm\)43.46       &\(\pm\)9.60        &\(\pm\)0           &                       &\(\pm\)7.02    &\(\pm\)12.83   &\(\pm\)17.60       &\(\pm\)30.14   &\(\pm\)10.63      \\
\hline
\mr{Grammar VAE*}                        &55.82              &5.06               &\dcb{}100.00       &\mr{94.64}             &\dcb{}99.21   &80.99         &50.24              &26.75          &25.42              \\
                                        &\(\pm\)49.66       &\(\pm\)22.99       &\(\pm\)0           &                       &\(\pm\)4.47    &\(\pm\)11.40   &\(\pm\)33.65       &\(\pm\)33.14   &\(\pm\)14.91       \\
\hline
\mr{Syntax Directed VAE*}                &\dcb{}77.38        &19.00              &\dcb{}100.00       &\dcb{}                 &93.56          &77.84          &55.94             &\dcb{}14.46          &39.45              \\
                                        &\(\pm\)41.84       &\(\pm\)39.23       &\(\pm\)0           &\dcb{}\mrm{100.00}     &\(\pm\)18.50    &\(\pm\)19.76   &\(\pm\)27.51       &\(\pm\)24.14   &\(\pm\)20.98       \\
\hline
\mr{Graph VAE}                          &0.27               &62.63              &\dcb{}100.00       &\mr{99.99}             &71.49          &90.68          &80.79              &28.07          &45.96              \\
                                        &\(\pm\)4.58        &\(\pm\)48.38       &\(\pm\)0           &                       &\(\pm\)25.36   &\(\pm\)11.71   &\(\pm\)17.33       &\(\pm\)20.14   &\(\pm\)18.69        \\
%\hline -- before cutting
%\mr{Graph VAE}                          &0.21               &49.13              &\dcb{}100.00       &\dcb{}                 &73.21          &\dcb{}88.02          &\dcb{}77.47        &31.00          &47.89              \\
%                                        &\(\pm\)4.58        &\(\pm\)49.99       &\(\pm\)0           &\dcb{}\mrm{100.00}     &\(\pm\)24.69    &\(\pm\)13.27  &\(\pm\)17.96       &\(\pm\)20.34   &\(\pm\)18.88        \\
\hline
\mr{Regularized GVAE}                   &0.01               &86.47              &\dcb{}100.00       &\mr{90.33}              &97.88          &\dcb{}95.88    &\dcb{}94.42        &44.64     &34.41              \\
                                        &\(\pm\)0.77         &\(\pm\)34.21      &\(\pm\)0           &                       &\(\pm\)6.96     &\(\pm\)6.84    &\(\pm\)9.61        &\(\pm\)25.14    &\(\pm\)13.26        \\
%\hline -- before cutting
%\mr{Regularized GVAE}                   &0.01               &99.66              &\dcb{}100.00       &\mr{1.78}              &\dcb{}99.99   &75.11          &35.26              &\dcb{}0.97     &36.55              \\
%                                        &\(\pm\)0.77         &\(\pm\)2.00        &\(\pm\)0           &                       &\(\pm\)0.17    &\(\pm\)1.16    &\(\pm\)9.20        &\(\pm\)6.86    &\(\pm\)2.30        \\
%\hline
%\mr{Regularized GVAE full}              &0.09               &50.46              &\dcb{}100.00       &\dcb{}                 &76.92          &\dcb{}90.63    &\dcb{}79.54        &29.14           &47.63              \\
%                                        &\(\pm\)3.00         &\(\pm\)50.00       &\(\pm\)0           &\dcb{}\mrm{100.00}     &\(\pm\)21.90   &\(\pm\)11.84   &\(\pm\)17.23       &\(\pm\)20.25    &\(\pm\)18.66        \\
\hline
\mr{Junction Tree VAE*}                  &50.23              &99.59              &99.98              &\mr{99.75}             &32.96          &52.20          &48.06              &44.74          &\dcb{}75.05        \\
                                        &\(\pm\)50.00       &\(\pm\)6.35        &\(\pm\)1.23        &                       &\(\pm\)21.78   &\(\pm\)17.12   &\(\pm\)18.48       &\(\pm\)24.39   &\(\pm\)13.40        \\
\hline
% version with lambda
%\mr{Constrained GVAE \(\lambda\)=0.3*}  &0.27               &\dcb{}100.00       &\dcb{}100.00       &\mr{99.84}             &62.13          &78.93          &56.70              &18.91          &62.89              \\
%                                        &\(\pm\)5.19        &\(\pm\)0            &\(\pm\)0          &                       &\(\pm\)24.79   &\(\pm\)15.07   &\(\pm\)20.53       &\(\pm\)22.05   &\(\pm\)17.62        \\
\mr{Constrained GVAE*}                  &0.35               &\dcb{}100.00       &\dcb{}100.00       &\mr{99.92}             &65.98          &81.38          &57.76              &16.25          &65.14              \\
                                        &\(\pm\)5.91        &\(\pm\)0            &\(\pm\)0          &                       &\(\pm\)22.78   &\(\pm\)15.98   &\(\pm\)20.04       &\(\pm\)21.63   &\(\pm\)16.39        \\
\thickhline
\dcf{}                                  &\multicolumn{4}{c}{}                                                               &               & 42.08         & 56.11             & 55.95         & 73.18             \\
\dcf{}\mrm{Properties' Scores for Dataset ZINC}                &\multicolumn{4}{c}{}                                        &               &\(\pm\)18.37   &\(\pm\)17.44       &\(\pm\)22.90   &\(\pm\)13.86        \\
\hline
\end{tabular}
\caption{Average and standard deviation of different metrics computed on the QM9 and ZINC dataset. The symbol '*' denotes models where we used values for the parameters tuned by the authors; entries with blue background highlight the best score; up and down arrows denote whether the metric should be maximized ($\uparrow$) or minimized ($\downarrow$). \label{tab:results}}
\end{table}
\end{landscape}
}

% Before cutting
%\caption{Results obtained on the QM9 and ZINC datasets. For each dataset: the symbol '*' denotes models where we used values for the parameters tuned by the authors for that dataset; entries with blue background highlight the best score obtained for each metric; up and down arrows in front of metrics name denote whether the metric should be maximized ($\uparrow$) or minimized ($\downarrow$). Average and standard deviation (where applicable) computed on the generated molecules are reported. Property scores for each dataset are reported as well. \label{tab:results}}
Tables~\ref{tab:results} reports the average and standard deviation of the experimental assessments (using the procedure in Section~\ref{sec:processesandmetrics}) on the models presented in Section~\ref{sec:models}.
We have shown the strengths and weaknesses of existing models and for the first time we have also reported the standard deviation.
%obtained by the models evaluation rounded to two decimal places.
The last line of each table reports the properties' scores obtained from the molecules in the datasets. The models should learn the distribution of the input data and for this reason it is expected that each model scores reflect those in the datasets.
Note that some of the results we report are slightly different from the ones reported in the original papers. Depending on the cases, this is due to the different evaluation procedure, high variance in the results,  or bugs in the original code that we fixed.
%Graph VAE model code is the same used for the Regularized Graph VAE model code applying only the standard VAE loss function e.g. it doesn't consider others constraints.
%This allows us to compare the two models, showing the effectiveness of the added constraints in the regularized loss function model.
Character VAE%, as anticipated in section \ref{sec:models}, 
~shows low \emph{validity} and \emph{reconstruction} in both  datasets.
%since not all SMILES strings it generates are actually molecules.
On the contrary, since a small modification to the SMILES string can correspond to a large modification of the molecular structure, it presents a high value of \emph{uniqueness} and \emph{novelty}.
%TODO forse tagliare?
%Unlike the mean of 44.6 reported by the authors, we obtain a lower \emph{reconstruction} in the ZINC dataset. This is to be attributed to the different molecules generation process, in which the author encode the molecules 10 times and decode each of these points 100 times. In fact using the same process we obtain values consistent with the one of the authors. In turn the \emph{validity} increase in the QM9 dataset, while the \emph{reconstruction} decrease. This is probably due to the high variance.
%-----
Grammar VAE, compared to Character VAE,
%, thanks to the use of the context-free grammar that shrink the solution space generating only those strings that are valid according to the SMILES grammar, 
increases the \emph{reconstruction} and the \emph{validity} scores while maintaining a high \emph{novelty} and a high \emph{uniqueness} in both datasets.
However, in QM9 this model has a similar \emph{validity} value as  Character VAE, probably because of the observed very high variance.
%In the ZINC dataset the difference among these two model are more visible.
Syntax Directed VAE improves Grammar VAE results in both datasets, even though in the QM9 the \emph{reconstruction} value is a bit lower than the one of Grammar VAE. This is probably due to the fact that the model parameters are not tuned on the QM9 dataset, and again to the presence of high variance.
%TODO forse tagliare
%The \emph{validity} is lower than that reported by the authors (43.5) on the ZINC dataset and this is attributed to the different molecule generation process that samples 1000 points from the latent space and decode each of them 100 times.
%------
Regularized Graph VAE reaches similar results to Graph VAE in the QM9 dataset, presenting good \emph{validity} and \emph{uniqueness} values, but very low \emph{reconstruction} and only about 40\% of \emph{novelty}.
In the ZINC dataset, the differences w.r.t. Graph VAE model are more evident.
In fact Regularized Graph VAE presents higher \emph{validity} value and a lower \emph{uniqueness} value.
%------ before cutting
%In the ZINC dataset, the \emph{reconstruction} and the \emph{uniqueness} values are very low, indicating that the model generates almost the same molecules every time.   
%Moeover, the generated molecules are simple (carbon molecules with a low number of atoms), as also confirmed by the \emph{SAS} metric.  
%------
%In fact, the \emph{SAS} metric also indicates that this model generates molecules that are simple to synthesize.
%TODO tagliare
%We think that this is due to the fact that by adding the valences constraint in the loss function, the model learns to generate molecules formed by only few atoms since in this way it is much easier to respect the constraints on valences. 
%On the contrary it would be much more complex to respect all the valences using bigger molecules, making the loss function bigger.
Junction Tree VAE presents high \emph{validity}, \emph{novelty} and \emph{uniqueness} values, in both datasets, even if it is optimized only on  ZINC. %TODO tagliare
%The \emph{reconstruction} is lower from that one reported by the authors (76.7) but again this is due to the different process uses to generate the molecule and the high variance.In fact the authors encoded the molecule 10 times in the latent space as a point and decode each of them 10 times.
However, since it generates molecules using substructures extracted from the training set, it tends to present a lower value on the \emph{diversity} score.
Constrained Graph VAE presents high \emph{validity}, \emph{novelty} and \emph{uniqueness} values,
%that are consistent with those of the authors 
in both datasets.
Considering the \emph{reconstruction} values,  
%obtained from the new implemented function in both the dataset, it is clear that this model is not suitable for solving the \emph{reconstruction} task. Moreover, if we look to the different \emph{reconstruction} values in the two dataset, 
this model has troubles reconstructing the complex molecules in ZINC.
%This is probably due to the model molecules reconstruction algorithm which, starting from atoms initially unconnected sampled from the latent space, through a breath-first search tries to link them with a bond.
%\todo{L'autore non lo dice esplicitamente, ma dice: "as our current one-shot prediction of the adjacency tensor is most likely feasible only for graphs of small size. Magari riportiamo la stessa frase?"}
MolGAN is trained only on the QM9 dataset because, as reported from the authors, this model doesn't scale well with larger molecules. 
It presents good \emph{validity} and \emph{novelty} values, but low \emph{uniqueness}
%TODO tagliare
%, even if this is higher than the one reported by the author (2.9) in which the model was trained only on a 5k subset of training molecules. However this low values happen 
due to the problem of the collapse of the model that is often present in GAN models. 
Since the goal of the generation process is to find new molecules with certain properties for high-throughput screening, we argue that this model is not particularly suited for this task.
%
%does not lend itself well to the solution of this task.

Looking at the metrics measuring the chemical properties (NP, Sol.,SAS and QED), % exhibit by the molecules, 
it seems that Junction Tree VAE and Constrained Graph VAE are the models able to best capture the characteristics of both datasets, even if the latter model tends to generate molecules that are simple to synthesize.
Although in the QM9 dataset
the considered models do not differ too much in the  measures of chemical properties,
%exhidataset the models properties results do not differ too much among them,
%in the ZINC dataset 
in ZINC these differences are more evident.
\section{Conclusions}
\label{sec:conclusions}
%\todo{forse è da rimuovere la virgola prima dell'and?}
Deep learning models for the generation of molecules are still in their infancy and they are not properly compared to each other.
%is a problem that still leaves room for improvement.
In fact, different models are tested by the authors on different datasets, with different evaluation processes and metrics, making it difficult to objectively compare them.
In this paper, we have proposed a set of processes for the evaluation of existing models according to relevant metrics, for which we have reported experimental results on two commonly adopted datasets. To ease future comparisons, 
we publicly released the code used for the reported assessment.
%for encouraging their wide adoption.
%in the state-of-the-art, implement all functions for reading datasets and for the generation of the molecules when necessary.

% include tables

% ****************************************************************************
% BIBLIOGRAPHY AREA
% ****************************************************************************

%\clearpage
\begin{footnotesize}

\bibliographystyle{unsrt_drigoni.bst}
\bibliography{abbr,references}

\end{footnotesize}

% ****************************************************************************
% END OF BIBLIOGRAPHY AREA
% ****************************************************************************

\end{document}